\documentclass[10pt,twocolumn,letterpaper]{article}

\usepackage{cvpr}              
\usepackage[accsupp]{axessibility}

\usepackage{graphicx}
\usepackage{amsmath}
\usepackage{amssymb}
\usepackage{booktabs}

\usepackage{mathtools} 
\usepackage{amssymb,amsfonts,bm,bbold} 
\usepackage{systeme}
\usepackage{url}
\usepackage{nicefrac} 
\usepackage{multirow} 

\usepackage[pagebackref,breaklinks,colorlinks]{hyperref}

\usepackage[capitalize]{cleveref}
\crefname{section}{Sec.}{Secs.}
\Crefname{section}{Section}{Sections}
\Crefname{table}{Table}{Tables}
\crefname{table}{Tab.}{Tabs.}

\def\cvprPaperID{6648} 
\def\confName{CVPR}
\def\confYear{2022}

\makeatletter
\newcommand{\mypm}{\mathbin{\mathpalette\@mypm\relax}}
\newcommand{\@mypm}[2]{\ooalign{%
  \raisebox{.1\height}{$#1+$}\cr
  \smash{\raisebox{-.6\height}{$#1-$}}\cr}}
\makeatother

\newcommand\norm[1]{\left\lVert#1\right\rVert}

\newcommand{\bo}[1]{\bm{#1}}

\def\vtheta{{\bm{\theta}}}

\def\vlambda{{\bm{\lambda}}}

\def\vg{{\bm{g}}}

\def\vp{{\bm{p}}}

\def\vr{{\bm{r}}}

\def\vx{{\bm{x}}}

\def\vz{{\bm{z}}}

\DeclareMathOperator*{\argmin}{arg\,min}

\begin{document}

\title{Recurrence without Recurrence: \\Stable Video Landmark Detection with Deep Equilibrium Models}

\author{Paul Micaelli\\
University of Edinburgh\\
{\tt\small paul.micaelli@ed.ac.uk}
\and
Arash Vahdat\\
NVIDIA\\
{\tt\small avahdat@nvidia.com}
\and
Hongxu Yin\\
NVIDIA\\
{\tt\small dannyy@nvidia.com}
\and
Jan Kautz\\
NVIDIA\\
{\tt\small jkautz@nvidia.com}
\and
Pavlo Molchanov\\
NVIDIA\\
{\tt\small pmolchanov@nvidia.com}
}
\maketitle

\newcommand\paul[1]{\textcolor{blue}{[Paul: #1]}}
\newcommand\JK[1]{\textcolor{magenta}{[JK: #1]}}
\newcommand\yin[1]{\textcolor{cyan}{[yin: #1]}}

\begin{abstract}

Cascaded computation, whereby predictions are recurrently refined over several stages, has been a persistent theme throughout the development of landmark detection models. In this work, we show that the recently proposed Deep Equilibrium Model (DEQ) can be naturally adapted to this form of computation. Our Landmark DEQ (LDEQ) achieves state-of-the-art performance on the challenging WFLW facial landmark dataset, reaching $3.92$ NME with fewer parameters and a training memory cost of $\mathcal{O}(1)$ in the number of recurrent modules. Furthermore, we show that DEQs are particularly suited for landmark detection in videos. In this setting, it is typical to train on still images due to the lack of labelled videos. This can lead to a ``flickering'' effect at inference time on video, whereby a model can rapidly oscillate between different plausible solutions across consecutive frames. By rephrasing DEQs as a constrained optimization, we emulate recurrence at inference time, despite not having access to temporal data at training time. This Recurrence without Recurrence (RwR) paradigm helps in reducing landmark flicker, which we demonstrate by introducing a new metric, normalized mean flicker (NMF), and contributing a new facial landmark video dataset (WFLW-V) targeting landmark uncertainty. On the WFLW-V hard subset made up of $500$ videos, our LDEQ with RwR improves the NME and NMF by $10$ and $13\%$ respectively, compared to the strongest previously published model using a hand-tuned conventional filter. 

\end{abstract}

\section{Introduction}
\label{sec:intro}

The field of facial landmark detection has been fueled by important applications such as face recognition \cite{Masi2018DeepFaceRecognitionSurvey1, Wang2018DeepFaceRecognitionSurvey2}, facial expression recognition \cite{Shan2021DeepFacialExpressionRecognitionSurvey, Jung2015FaceExpressionLandmarkTrajectory, Kim2017MultiModalEmotionRecognition, Yan2016MultiClueFusion, Hasani2017EnhancedDeep3DCNNs}, and face alignment \cite{Xiangyu2015HighFidelityPoseNormalization, Mallikarjun2015EfficientFaceFrontalization, Zhang2013PoseInvariantRecognition}. Early approaches to landmark detection relied on a statistical model of the global face appearance and shape \cite{Edwards1998InterpretingFaceImagesWithAAM, Cootes2011AAM, Cristinacce2006ConstrainedLocalModels}, but this was then superseded by deep learning regression models \cite{Wayne2018LookAtBoundary, Feng2018WingLoss, Wan2020MultiOrderHG, Valle2018EnsembleOfRegressionTrees, Dapogny2019DeCaFA, Sun2019HRNet, Qian2019AggregationViaSeparation, Kumar2020Luvli, Wang2019AdaptiveWingLoss, Lin2021StructureCoherent,  Li2020StructuredLandmarkDetection, Huang2021ADNet, Xia2022SLPT}. Both traditional and modern approaches have relied upon cascaded computation, an approach which starts with an initial guess of the landmarks and iteratively produces corrected landmarks which match the input face more finely. These iterations typically increase the training memory cost linearly, and do not have an obvious stopping criteria. To solve these issues, we adapt the recently proposed Deep Equilibrium Model \cite{Bai2019DEQs, Bai2020MDEQs, Bai2021DEQJacobianRegularization} to the setting of landmark detection. Our Landmark DEQ (LDEQ) achieves state-of-the-art performance on the WFLW dataset, while enjoying a natural stopping criteria and a memory cost that is constant in the number of cascaded iterations. 

\begin{figure*}[t!]
  \centering
   \includegraphics[width=\linewidth]{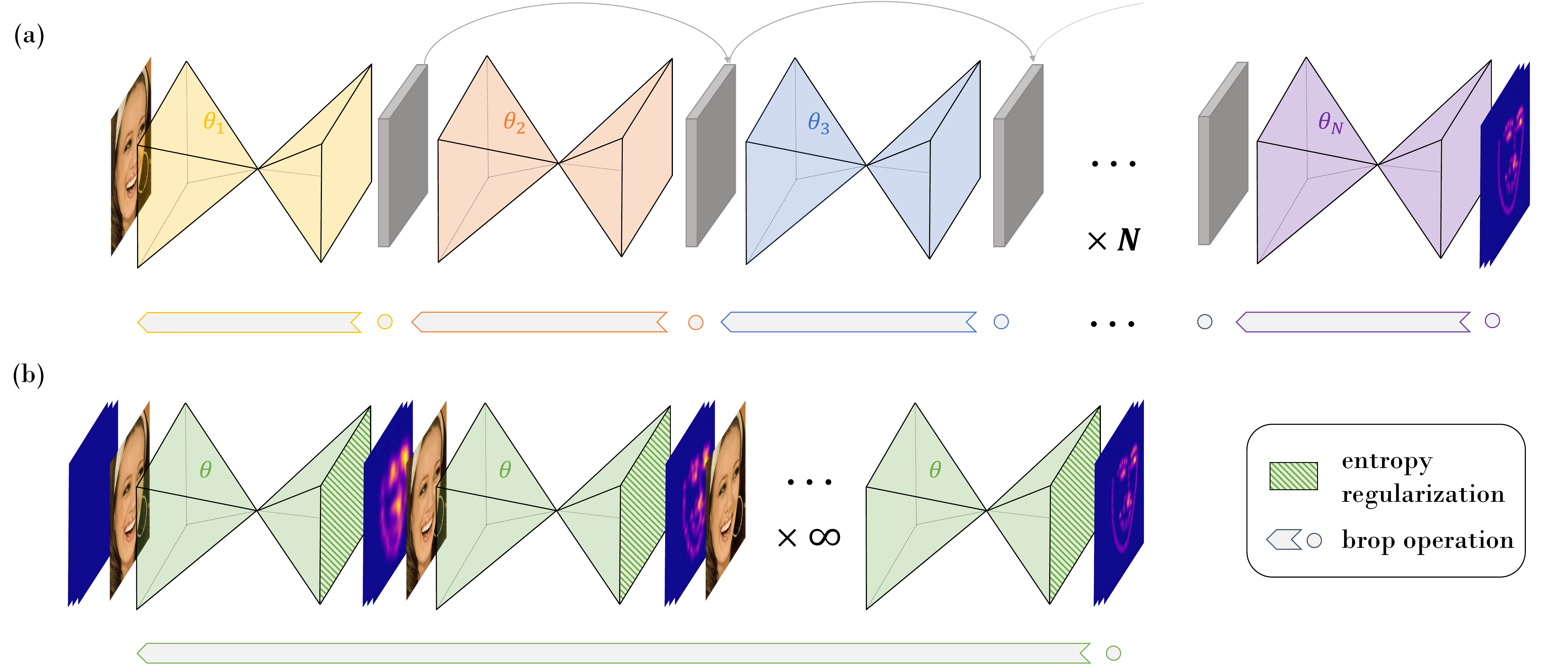}
   \caption{(a) Common Stacked-Hourglass architecture \cite{Newell2016StackedHourglass}, whereby each cascaded stage increases the memory cost and the number of backpropagation operations. (b) Our LDEQ model, which adapts an equilibrium model \cite{Bai2019DEQs} to the landmark detection setting, enjoying a constant memory cost with the number of refining stages. At each stage, we compute landmark probability heatmaps, and encourage convergence to an equilibrium by lowering their entropy. We release our code here: {\url{https://github.com/polo5/LDEQ_RwR}}.} 
   \label{fig:deq vs stacked hourglass}
\end{figure*}

Furthermore, we explore the benefits of DEQs in landmark detection from facial videos. Since obtaining landmark annotation for videos is notoriously expensive, models are virtually always trained on still images and applied frame-wise on videos at inference time. When a sequence of frames have ambiguous landmarks (\eg, occluded faces or motion blur), this leads to flickering landmarks, which rapidly oscillate between different possible configurations across consecutive frames. This poor temporal coherence is particularly problematic in applications where high precision is required, which is typically the case for facial landmarks. These applications include face transforms \cite{MediaPipeGoogleBlog, MediaPipeGoogleBlog3D},  face reenactment \cite{Zhang2019FReeNetFaceReenactment}, video emotion recognition \cite{Jung2015FaceExpressionLandmarkTrajectory , Kim2017MultiModalEmotionRecognition, Yan2016MultiClueFusion, Hasani2017EnhancedDeep3DCNNs}, movie dubbing \cite{Garrido2015DubbingAudioTracks} or tiredness monitoring \cite{Jabbar2020DriverTiredness}. We propose to modify the DEQ objective at inference time to include a new recurrent loss term that encourages temporal coherence. We measure this improvement on our new WFLW-Video dataset (WFLW-V), demonstrating superiority over traditional filters, which typically reduce flickering at the cost of reducing landmark accuracy. 

\section{Related work}
\label{sec:related_work}

Deep Equilibrium Models \cite{Bai2019DEQs} are part of the family of implicit models, which learn implicit functions such as the solution to an ODE \cite{Ricky2018NeuralODEs, Dupont2019AugmentedNeuralODEs}, or the solution to an optimization problem \cite{Amos2017Optnet, Djolonga2017DifferentiableLearningSubmodularModels, Wang2019Satnet}. These functions are called implicit in the sense that the output cannot be written as a function of the input explicitly. In the case of DEQs, the output is the root of a function, and the model learned is agnostic to the root solver used. Early DEQ models were too slow to be competitive, and much work since has focused on better architecture design \cite{Bai2020MDEQs} and faster root solving \cite{Bai2021DEQJacobianRegularization, Fung2021JacobianFreeBackprop, Zhengyang2021OnTrainingImplicitModels}. Since the vanilla formulation of DEQs does not guarantee convergence or uniqueness of an equilibrium, another branch of research has focused on providing convergence guarantees \cite{Winston2020MonotoneOperatorDEQs, Revay2020LipschitzBoundedDEQs, Pabbaraju2021EstimatingLipschitzConstantsOfDEQs}, which usually comes at the cost of a performance drop. Most similar to our work is the recent use of DEQs for videos in the context of optical flow estimation \cite{Bai2022DeqFlow}, where slightly better performance than LSTM-based models was observed.

A common theme throughout the development of landmark detection models has been the idea of cascaded compute, which has repeatedly enjoyed a better performance compared to single stage models \cite{Wang2018DeepFaceRecognitionSurvey2}. This is true in traditional models like Cascaded Pose Regression (CPR) \cite{Dollar2010CascadedPoseRegression, Xudong2012ExplicitShapeRegression, Xiong2013SupervisedDescentMethod, Sun2013ConvolutionalCascade, Asthana2014IncrementalFaceAlignment}, but also in most modern landmark detection models \cite{YangStackedHourglassForFacialLandmark, Wayne2018LookAtBoundary, Wan2020MultiOrderHG, Kumar2020Luvli, Wang2019AdaptiveWingLoss, Huang2021ADNet, Lan2021HIH}, which usually rely on the Stacked-Hourglass backbone \cite{Newell2016StackedHourglass}, or an RNN structure \cite{Trigeorgis2016MnemonicDescentMethod, Hanjiang2018DeepRecurrentRegression}. In contrast to these methods, our DEQ-based model can directly solve for an infinite number of refinement stages at a constant memory cost and without gradient degradation. This is done by relying on the implicit function theorem, as opposed to tracking each forward operation in autograd. Furthermore, our model naturally supports adaptive compute, in the sense that the number of cascaded stages will be determined by how hard finding an equilibrium is for a specific input, while the number of stages in, say, the Stacked-Hourglass backbone, must be constant at all times.

Our recurrence without recurrence approach is most closely related to test time adaption methods. These have been most commonly developed in the context of domain adaptation \cite{Sun2020TestTimeTraining, Wang2021TestTimeAdaptationByEntropyMinimization}, reinforcement learning \cite{Hansen2021SelfSupervisedPolicy}, meta-learning \cite{Zhang2021AdaptiveRiskMinimization}, generative models \cite{Jiang2021HandObjectGeneration, Mu2021ArticulatedShapeRepresentation}, pose estimation \cite{Li2021TestTimePoseEstimation} or super resolution \cite{Assaf2018ZeroShotSuperResolution}. Typically, the methods above finetune a trained model at test time using a form of self-supervision. In contrast, our model doesn't need to be fine-tuned at test time: since DEQs solve for an objective function in the forward pass, this objective is simply modified at test time.

\section{DEQs for landmark detection}
\label{sec: deq formulation}

Consider learning a landmark detection model $F$ parameterized by $\vtheta$, which maps an input image $\vx$ to landmarks $\vz$. Instead of directly having $\vz$ as 2D landmarks, it is common for $\vz$ to represent $L$ heatmaps of size $D \times D$, where $D$ is a hyperparameter (usually $64$) and $L$ is the number of landmarks to learn. While typical machine learning models can explicitly write down the function $\vz = F(\vx; \vtheta)$, in the DEQ approach this function is expressed implicitly by requiring its output to be a fixed point of another function $f(\vz, \vx; \vtheta)$:
\begin{equation}
 F : \vx \rightarrow \vz^* ~~~ \text{s.t.}~~~ \vz^* = f(\vz^*, \vx; \vtheta)  
\end{equation}
where $\vz^*$ denotes the fixed point, or equivalently the root of $g(\vx, \vz; \vtheta) = f(\vz, \vx; \vtheta) - \vz$. The function $f$ must have inputs and outputs of similar shape, but beyond this restriction there is still limited understanding of its desired properties in machine learning. For simplicity, we build $f$ from the ubiquitous hourglass module $h$ (similar to a Unet): 
\begin{equation}
f(\vz, \vx; \vtheta) = \sigma ( h([\vx,\vz]; \vtheta) )
\end{equation}
where $h$ inputs the concatenation of $\vx$ and $\vz$, and $\sigma$ is a normalization function. For clarity, we omit from our notation that image $x$ is downsampled with a few convolutions to match the shape of $\vz$. 

To evaluate $f$ in the forward pass, we must solve for its fixed point $\vz^*$. When $f$ is a contraction mapping, $f \circ f \circ f \circ \cdots \circ f \circ \vz^{(0)}$ converges to a unique $\vz^*$ for any initial heatmap $\vz^{(0)}$. In practice, it is neither tractable or helpful to directly take an infinite number of fixed point iteration steps. Instead, it is common to achieve the same result by leveraging quasi Newtonian solvers like Broyden's method \cite{Broyden1965ACO} or Anderson acceleration \cite{Anderson1965IterativeProcedure}, which find $\vz^*$ in fewer iterations. Similarly to the original DEQ model, we use $\vz^{(0)} = \bo{0}$ when training our LDEQ on still images.

Guaranteeing the existence of a unique fixed point by enforcing contraction restrictions on $f$ is cumbersome, and better performance can often be obtained by relying on regularization heuristics that are conducive to convergence, such as weight normalization and variational dropout \cite{Bai2020MDEQs}. In our landmark model, we did not find these tricks helpful, and instead used a simple normalization of the heatmaps to $[0,1]$ at each refining stage:
\begin{equation}
\sigma(\vz) = \exp \left(\frac{\vz - \max(\vz)}{T}\right)
\end{equation}
where $T$ is a temperature hyperparameter. This layer also acts as an entropy regularizer, since it induces low-entropy (``peaked'') heatmaps, which we found to improve convergence of root solvers. 

We contrast our model in \cref{fig:deq vs stacked hourglass} to the popular Stacked-Hourglass backbone \cite{Newell2016StackedHourglass}. Contrary to this model, our DEQ-based model uses a single hourglass module which updates $\vz$ until an equilibrium is found. The last predicted heatmaps, $\vz^*$, are converted into 2D landmark points $\hat{\vp} = \Phi(\vz^*)$ using the softargmax function $\Phi(\vz)$ proposed in \cite{Luvizon2017SoftArgmax}. These points are trained to match the ground truth $\vp$, and so the DEQ landmark training problem can be seen as a constrained optimization:
\begin{equation}
\begin{aligned}
    \vtheta^* &= \argmin_{\vtheta} ~~\mathcal{L}_{\text{MSE}}~(\Phi(\vz^*), \vp) \\
    & ~~~\text{s.t.}~~~ \vz^* = f(\vz^*, \vx; \vtheta)
\end{aligned}
\label{eq: constrained deq main formulation}
\end{equation}
To differentiate our loss function $\mathcal{L}_{\text{MSE}}$ through this root solving process, the implicit function theorem is used \cite{Bai2019DEQs} to derive
\begin{equation}
\frac{\partial \mathcal{L}_{\text{MSE}}} {\partial \vtheta} = - \frac{\partial \mathcal{L}_{\text{MSE}}} {\partial \vz^*}  \mathcal{J}^{-1}_{\vg \vz^*} \frac{\partial f(\vz^*, \vx;\vtheta) }{\partial \vtheta}
\end{equation}
where the first two terms on the RHS can be expressed as the solution to a fixed point problem as well. Solving for this root in the backward pass means that we do not need to compute or store the expensive inverse Jacobian term $\mathcal{J}^{-1}_{\vg \vz^*}$ directly \cite{Bai2019DEQs, DeepImplicitLayerTutorial}. Importantly, this backward-pass computation only depends on $\vz^*$ and doesn't depend on the operations done in the forward pass to reach an equilibrium. This means that these operations do not need to be tracked by autograd, and therefore that training requires a memory cost of O(1), despite differentiating through a potentially infinite recurrence.

\section{Recurrence without recurrence}

\begin{figure*}[t!]
  \centering
   \includegraphics[width=\linewidth]{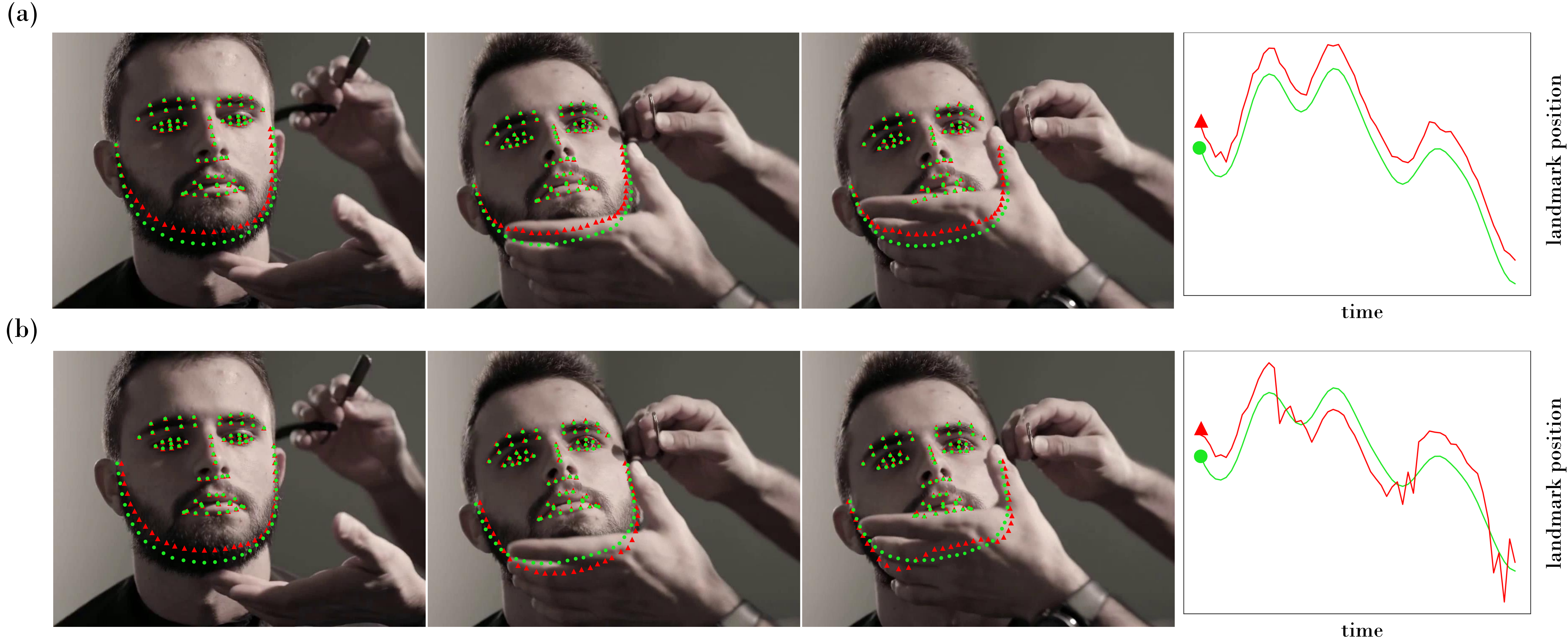}
   \caption{Model predictions (\textcolor{red}{$\blacktriangle$}) and ground truth landmarks (\textcolor{green}{\textbullet}) for (a) a model with high temporal coherence and (b) a model with the same accuracy but exhibiting a worse temporal coherence, due to ambiguity at the chin. This causes flickering around the ground truth, as illustrated in the landmark trajectory (right). This flickering is common when video frames are evaluated individually, as opposed to recurrently.} 
   \label{fig:high flicker vs low flicker}
\end{figure*}

Low temporal coherence (i.e. a large amount of flicker) is illustrated in \cref{fig:high flicker vs low flicker}. This can be a nuisance for many applications that require consistently precise landmarks for video. In this section, we describe how our LDEQ model can address this challenge by enabling recurrence at test time without recurrence at training time (RwR).

Recall that in the DEQ formulation of \cref{sec: deq formulation}, there is no guarantee that a unique fixed point solution exists. This can be a limitation for some applications, and DEQ variants have been proposed to allow provably unique solutions at the cost of additional model complexity \cite{Winston2020MonotoneOperatorDEQs, Pabbaraju2021EstimatingLipConstantsOfMonDEQs}. In this work, we instead propose a new paradigm: we leverage the potentially large solution space of DEQs after training to allow for some additional objective at inference time. This new objective is used to disambiguate which fixed point of $f(\vz, \vx;\vtheta^*)$ is found, in light of extra information present at test time. We demonstrate this approach for the specific application of training a landmark model on face images and evaluating it on videos. In this case, DEQs allow us to include a recurrent loss term at inference time, which isn't achievable with conventional architectures.

Let $f(\vz,\vx;\vtheta^*)$ be our DEQ model trained as per the formulation in \cref{eq: constrained deq main formulation}. We would now like to do inference on a video of $N$ frames $\vx_1, \vx_2, \cdots, \vx_N$. Consider that a given frame $\vx_n$ has a corresponding set of fixed points $\mathcal{Z}^*_n = \{ \vz ~~ \text{s.t.} ~~ f(\vz, \vx_n; \vtheta^*) = \vz \}$, representing plausible landmark heatmaps. If we select some $\vz_n^* \in \mathcal{Z}^*_n$ at random for each frame $n$, the corresponding heatmaps $\vz^*_1, \vz^*_2, \cdots, \vz^*_N$ often exhibit some flickering artefacts, whereby landmarks rapidly change across contiguous frames (see \cref{fig:high flicker vs low flicker}). We propose to address this issue by choosing the fixed point at frame $n$ that is closest to the fixed point at frame $n-1$. This can be expressed by the following constrained optimization:%
\begin{subequations}
\begin{align}
\vz^*_n &= \argmin_{\vz} \norm{\vz - \vz^*_{n-1}}_2^2 \label{eq:constrained reg opt upper} \\
& ~~~ \text{s.t.} ~~ f(\vz, \vx_n; \vtheta^*) = \vz \label{eq:constrained reg opt lower}
\end{align}
\end{subequations}
The problem above is equivalent to solving for the saddle point of a Lagrangian as follows:
\begin{equation}
    \min_{\vz} \max_{\vlambda} \norm{\vz - \vz^*_{n-1}}_2^2 + \vlambda^T (f(\vz, \vx_n; \vtheta^*) - \vz)
\end{equation}
where $\vlambda$ are Lagrange multipliers. Effectively, we are using the set of fixed points $\mathcal{Z}^*_n$ in \cref{eq:constrained reg opt lower} as the trust region for the objective in \cref{eq:constrained reg opt upper}. In practice, adversarial optimization is notoriously unstable, as is typically observed in the context of GANs \cite{Goodfellow2014GAN, Arjovsky2014WassersteinGANs, Srivastava2017VeeGAN}. Furthermore, this objective breaks down if $\mathcal{Z}^*_n = \emptyset$ for any $\vx_n$. We can remedy both of these problems by relaxing the inference time optimization problem to:
\begin{equation}
\min_{\vz} \norm{f(\vz, \vx_n; \vtheta^*) - \vz}_2^2 + \frac{\alpha}{2} \norm{\vz - \vz^*_{n-1}}_2^2
\label{eq: relaxed RwR minimization}
\end{equation}
where $\alpha$ is a hyperparameter that trades off fixed-point solver error vs.\ the shift in heatmaps across two consecutive frames. This objective can be more readily tackled with Newtonian optimizers like L-BFGS \cite{Liu1989LBFGS}. When doing so, our DEQ at inference time can be described as a form of OptNet \cite{Amos2017Optnet}, albeit without any of the practical limitations (\eg, quadratic programs) related to making gradient calculations cheap. 

\begin{figure*}[t!]
  \centering
   \includegraphics[width=0.93\linewidth]{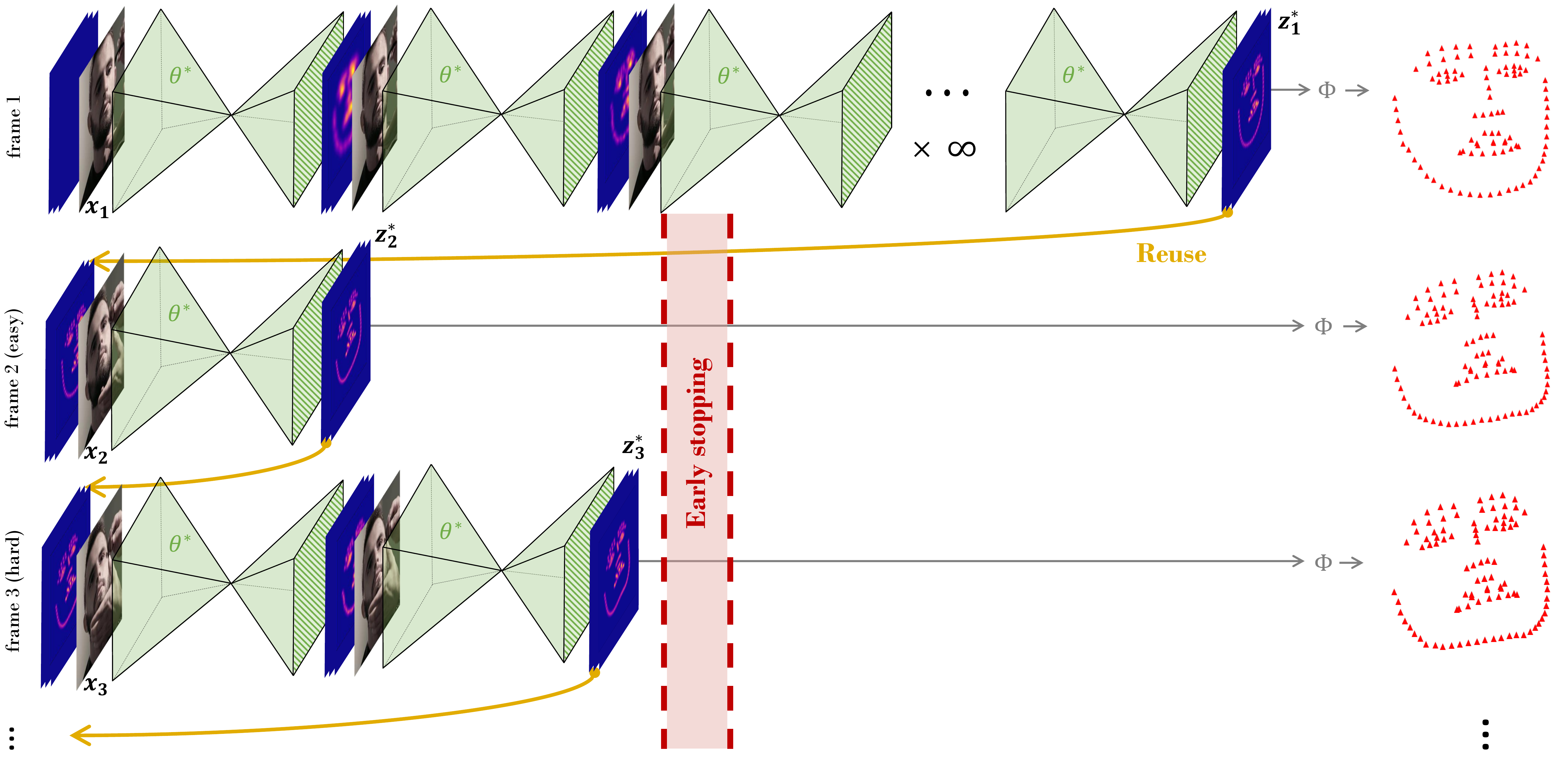}
   \caption{Recurrence without recurrence (RwR) on video data, from an LDEQ that was trained on still images. We use the initialization $\vz_1^{(0)} = \bo{0}$ for the first frame, and then reuse $\vz_n^{(0)} = \vz_{n-1}^*$ for $n > 1$. Combined with early stopping, this is equivalent to regularizing the fixed point $\vz_n$ so that it is more temporally coherent with all its predecessors.} 
   \label{fig: RwR}
\end{figure*}

Converting our root solving problem into an optimization problem during the forward pass of each frame can significantly increases inference time. Thankfully, the objective in \cref{eq: relaxed RwR minimization} can also be solved by using root solvers. First, note that it is equivalent to finding the MAP estimate given a log likelihood and prior: 
\begin{subequations}
\begin{align}
\log p(\vx_n | \vz; \vtheta^*) &\propto - \norm{f(\vz, \vx_n; \vtheta^*) - \vz}_2^2 \label{eq:likelihood} \\
p(\vz) &= \mathcal{N}(\vz; \vz_{n-1}, \alpha^{-1} \mathbb{1}) \label{eq:prior}
\end{align}
\end{subequations}
It has been demonstrated in various settings that this prior, when centered on the initialization to a search algorithm, can be implemented by early stopping this algorithm \cite{Sjoberg1992Overtraining, Bishop1995RegularizationComplexityControl, Santos1996EquivalenceRegAndTruncation, Grant2018RecastingMAMLasHierBayes}. As such, we can approximate the solution to Eq.~(\ref{eq:constrained reg opt upper}-\ref{eq:constrained reg opt lower}) by simply initializing the root solver with $\vz_n^{(0)} = \vz_{n-1}^*$ (``reuse'') and imposing a hard limit on the number of steps that it can take (``early stopping''). We call this approach Recurrence without Recurrence (RwR), and illustrate it in \cref{fig: RwR}.

\section{Video landmark coherence}

In this section we describe the metric (NMF) and the dataset (WFLW-V) that we contribute to measure the amount of temporal coherence in landmark videos. These are later used to benchmark the performance of our RwR paradigm against alternatives in \cref{subsec: exp-landmark coherence}. 

\subsection{NMF: a metric to track temporal coherence}

The performance of a landmark detection model is typically measured with a single metric called the Normalized Mean Error (NME). Consider a video sequence of $N$ frames, each containing $L$ ground truth landmarks. A single landmark point is a 2D vector denoted $\vp_{n,l} \in \mathcal{R}^2$, where $n$ and $l$ are the frame and landmark index respectively. Let $\vr_{n,l} = \vp_{n,l} - \bm{\hat{p}}_{n,l}$ be the residual vector between ground truth landmarks and predicted landmarks $\bm{\hat{p}}_{n,l}$. The NME simply averages the $\ell_2$ norm of this residual across all landmarks and all frames:
\begin{subequations}
\begin{align}
\text{NME}_{n} &= \frac{1}{L} \sum_{l=1}^{L} \frac{\norm{\vr_{n,l}}}{d_0} \\
\text{NME} &= \frac{1}{N} \sum_{n=1}^N \text{NME}_{n}
\label{eq: NME}
\end{align}
\end{subequations}
Here $d_0$ is usually the inter-ocular distance, and aims to make the NME better correlate with the human perception of landmark error. We argue that this metric alone is insufficient to measure the performance of a landmark detector in videos. In \cref{fig:high flicker vs low flicker} we show two models of equal NME but different coherence in time, with one model exhibiting \textit{flickering} between plausible hypotheses when uncertain. This flickering is a nuisance for many applications, and yet is not captured by the NME. This is in contrast to random noise (\textit{jitter}) which is unstructured and already reflected in the NME metric. 

To measure temporal coherence, we propose a new metric called the Normalized Mean Flicker (NMF). We design this metric to be orthogonal to the NME, by making it agnostic to the magnitude of $\vr_{n,l}$, and only focusing on the change of $\vr_{n,l}$ across consecutive frames:
\begin{subequations}
\begin{align}
\text{NMF}_{n} &= \sqrt{\frac{1}{L} \sum_{l=1}^{L} \frac{\norm{\vr_{n,l} - \vr_{n-1, l}}^2}{d^{2}_1}} \\
\text{NMF} &= \sqrt{\frac{1}{N} \sum_{n=2}^F \text{NMF}^{2}_{n}}
\label{eq: NMF}
\end{align}
\end{subequations}
We replace the means in the NME with a root mean square to better represent the human perception of flicker. Indeed, this penalizes a short sudden changes in $\vr_{n,l}$ compared to the same change smoothed out in time and space. The value $d_1^2$ is chosen to be the face area. This prevents a long term issue with the NME score, namely the fact that large poses can have an artificially large NME due to having a small $d_0$.

\subsection{A new landmark video dataset: WFLW-V}

\begin{figure*}[ht!]
  \centering
   \includegraphics[width=0.95\linewidth]{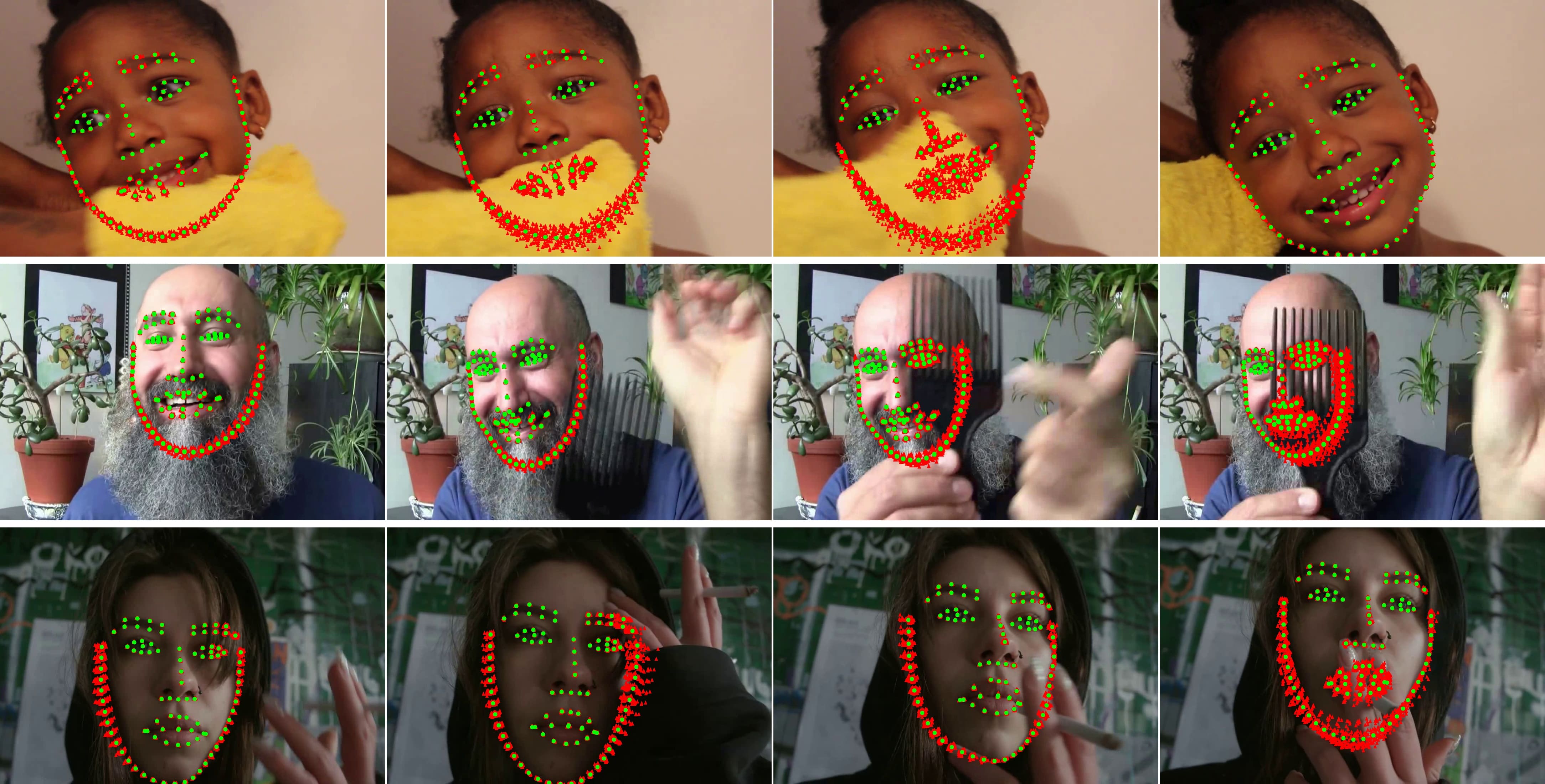}
   \caption{Example of ground truth labels (\textcolor{green}{\textbullet}) obtained semi-automatically from an ensemble of $45$ models (\textcolor{red}{$\blacktriangle$}). Ensembling these diverse models provides ground truth labels that do not flicker and can thus be used to measure flickering against. Note how landmark points with the most uncertainty are the most prone to having flickering predictions across consecutive frames.}
   \label{fig:examples of labelled videos}
\end{figure*}

Due to the tedious nature of producing landmark annotations for video data, there are few existing datasets for face landmark detection in videos. Shen \etal have proposed 300-VW \cite{Shen2015FirstFacialLandmarkTracking}, a dataset made up of $\sim 100$ videos using the 68-point landmark scheme from the 300-W dataset \cite{Sagonas2013Dataset300W}. Unfortunately, two major issues make this dataset unpopular: 1) it was labelled using fairly weak models from \cite{Chrysos2015DeformableFaceTracking} and \cite{Tzimiropoulos2015ProjectOutCascadedRegression} which results in many labelling errors and high flicker (see Appendix B), and 2) it only contains $100$ videos of little diversity, many of which being from the same speaker, or from different speakers in the same environment. Taken together, these two issues mean that the performance of modern models on 300-WV barely correlates with performance on real-world face videos.

We propose a new video dataset for facial landmarks: WFLW-Video (WFLW-V). It consists of $1000$ Youtube creative-commons videos (\ie, an order of magnitude more than its predecessor) and covers a wide range of people, expressions, poses, activities and background environments. Each video is $5$s in length. These videos were collected by targeting challenging faces, where ground truth landmarks are subject to uncertainty. The dataset contains two subsets, \textit{hard} and \textit{easy}, made up of $500$ videos each. This allows debugging temporal filters and smoothing techniques, whose optimal hyperparameters are often different for videos with little or a lot of flicker. This split was obtained by scraping $2000$ videos, and selecting the top and bottom $500$ videos based on the variance of predictions in the ensemble. To evaluate a landmark model on the WFLW-V dataset, we train it on the WFLW training set, and evaluate it on all WFLW-V videos. This pipeline best reflects the way large landmark models are trained in the industry, where labelled video data is scarce.
\begin{table*}[t!]
  \centering
  \resizebox{2.0\columnwidth}{!}{
  \begin{tabular}{@{}ll|c|cccccccc@{}}
    \toprule
    Method & & Params (M) & Full & Large poses & Expressions & Illumination & Makeup & Occlusion & Blur \\
    \midrule
    LAB \cite{Wayne2018LookAtBoundary} & \scriptsize{CVPR 2018} & 12.3 &  5.27 & 10.24 & 5.51 & 5.23 & 5.15 & 6.79 & 6.32 \\
    Wing \cite{Feng2018WingLoss} & \scriptsize{CVPR 2018} & 25 & 4.99 & 8.43 & 5.21 & 4.88 & 5.26 & 6.21 & 5.81 \\
    MHHN \cite{Wan2020MultiOrderHG} & \scriptsize{TIP 2020} & - & 4.77 & 9.31 & 4.79 & 4.72 & 4.59 & 6.17 & 5.82 \\
    DecaFA \cite{Dapogny2019DeCaFA} & \scriptsize{ICCV 2019} & 10 & 4.62 & 8.11 & 4.65 & 4.41 & 4.63 & 5.74 & 5.38 \\
    HRNet \cite{Sun2019HRNet} & \scriptsize{TPAMI 2020} & 9.7 & 4.60 & 7.94 & 4.85 & 4.55 & 4.29 & 5.44 & 5.42 \\
    AS \cite{Qian2019AggregationViaSeparation} & \scriptsize{ICCV 2019} & 35 & 4.39 & 8.42 & 4.68 & 4.24 & 4.37 & 5.60 & 4.86 \\
    LUVLI \cite{Kumar2020Luvli} & \scriptsize{CVPR 2020} & - & 4.37 & 7.56 & 4.77 & 4.30 & 4.33 & 5.29 & 4.94 \\
    AWing \cite{Wang2019AdaptiveWingLoss} & \scriptsize{ICCV 2019} & 24.2 & 4.36 & 7.38 & 4.58 & 4.32 & 4.27 & 5.19 & 4.96 \\
    SDFL \cite{Lin2021StructureCoherent} & \scriptsize{TIP 2021} & - & 4.35 & 7.42 & 4.63 & 4.29 & 4.22 & 5.19 & 5.08 \\
    SDL \cite{Li2020StructuredLandmarkDetection} & \scriptsize{ECCV 2020} & - & 4.21 & 7.36 & 4.49 & 4.12 & 4.05 & 4.98 & 4.82 \\
    ADNet \cite{Huang2021ADNet} & \scriptsize{ICCV 2021} & 13.4 & 4.14 & 6.96 & 4.38 & 4.09 & 4.05 & 5.06 & 4.79 \\
    SLPT \cite{Xia2022SLPT} & \scriptsize{CVPR 2022} & 19.5 &  4.12 & 6.99 & 4.37 & \textbf{4.02} & 4.03 & 5.01 & 4.79 \\
    HIH \cite{Lan2021HIH} & \scriptsize{~~~~~~~~~-} & 22.7 & 4.08 & 6.87 & 4.06 & 4.34 & 3.85 & 4.85 & 4.66 \\
    \midrule
    LDEQ (ours) & \scriptsize{~~~~~~~~~-} & 21.8 & \textbf{3.92} & \textbf{6.86} & \textbf{3.94} & 4.17 & \textbf{3.75} & \textbf{4.77} & \textbf{4.59} \\

    \bottomrule
  \end{tabular}
  }
  \caption{Performance of our model and previous state-of-the-art on the various WFLW subsets, using the NME metric ($\downarrow$) for comparison. Models using pre-training on other datasets have been excluded for fair comparison \cite{Bulat2021SubpixelHeatmaps, Zheng2021GeneralFacialRepresentationLearning, Yu2021HeatmapViaRandomizedRounding}.}
  \label{tab:WFLW performance NME}
  
\end{table*}

Contrary to the 68-landmark scheme of the 300-VW dataset, we label videos semi-automatically using the more challenging 98-landmark scheme from the WFLW dataset \cite{Wayne2018LookAtBoundary}, as it is considered the most relevant dataset for future research in face landmark detection \cite{Khabarlak2022FastFacialLandmarkDetectionSurvey1}. To produce ground truth labels, we train an ensemble of $45$ state-of-the-art models using a wide range of data augmentations of both the test and train set of WFLW (amounting to 10,000 images). We use a mix of large Unets, HRNets \cite{Sun2019HRNet} and HRFormers \cite{Yuan2021HRFormer} to promote landmark diversity. The heatmaps of these models are averaged to produce the ground truth heatmap, allowing uncertain models to weight less in the aggregated output. Ensembling models provides temporal coherence without the need for using hand-tuned filtering or smoothing, which are susceptible to misinterpreting signal for noise (\eg closing eye landmarks mimic high frequency noise). 

We provide examples of annotated videos in \cref{fig:examples of labelled videos}. Note that regions of ambiguity, such as occluded parts of the face, correspond to a higher variance in landmark predictions. While the ground truth for some frames may be subjective (\eg occluded mouth), having a temporally stable ``best guess'' is sufficient to measure flicker of individual models. We found that $45$ models in the ensemble was enough to provide a low error on the mean for all videos in WFLW-V. While we manually checked that our Oracle was highly accurate and coherent in time, note that it is completely impractical to use it for real-world applications due to its computational cost ($\sim 2B$ parameters). 

We checked each frame manually for errors, which were rare. When we found an error in annotation, we corrected it by removing inaccurate models from the ensemble for the frames affected, rather than re-labeling the frame manually. We found this approach to be faster and less prone to human subjectivity when dealing with ambiguous faces, such as occluded ones. Occlusion has been characterized as one of the main remaining challenges in modern landmark detection \cite{Wu2019FacialLandmarkDetectionSurvey2}, being most difficult in video data \cite{Shen2015FirstFacialLandmarkTracking}. Finally, the stability of our oracle also depends on the stability of the face bounding box detector. We found the most popular detector, MTCNN \cite{Zhang2017MTCNN} to be too jittery, and instead obtained stable detection by bootstrapping our oracle landmarks into the detector. More details about scraping and curating our dataset can be found in Appendix A.

\begin{table}[b!]
  \centering
  \resizebox{\columnwidth}{!}{
  \begin{tabular}{@{}c|c|ccccccc@{}}
    \toprule
    Metric & Method & Full & Pose & Exp. & Illum. & Mu. & Occ. & Blur \\
    \midrule
    \multirow{11}{*}{$\text{FR}_{10}$} & LAB & 7.56 & 28.83 & 6.37 & 6.73 & 7.77 & 13.72 & 10.74  \\
                          & HRNet & 4.64 & 23.01 & 3.50 & 4.72 & 2.43 & 8.29 & 6.34  \\
                          & AS & 4.08 & 18.10 & 4.46 & 2.72 & 4.37 & 7.74 & 4.40 \\
                          & LUVLi & 3.12 & 15.95 & 3.18 & 2.15 & 3.40 & 6.39 & 3.23 \\
                          & AWing & 2.84 & 13.50 & 2.23 & 2.58 & 2.91 & 5.98 & 3.75 \\
                          & SDFL & 2.72 & 12.88 & 1.59 & 2.58 & 2.43 & 5.71 & 3.62 \\
           ($\downarrow$) & SDL & 3.04 & 15.95 & 2.86 & 2.72 & 1.45 & 5.29 & 4.01 \\
                          & ADNet & 2.72 & 12.72 & 2.15 & 2.44 & 1.94 & 5.79 & 3.54 \\
                          & SLPT & 2.72 & 11.96 & 1.59 & \textbf{2.15} & 1.94 & 5.70 & 3.88 \\
                          & HIH & 2.60 & 12.88 & \textbf{1.27} & 2.43 & \textbf{1.45} & \textbf{5.16} & 3.10 \\
    \cmidrule{2-9}
    & LDEQ & \textbf{2.48} & \textbf{12.58} & 1.59 & 2.29 & 1.94 & 5.36 & \textbf{2.84}\\
    \midrule
    \multirow{11}{*}{$\text{AUC}_{10}$} & LAB & 0.532 & 0.235 & 0.495 & 0.543 & 0.539 & 0.449 & 0.463 \\
                           & HRNet & 0.524 & 0.251 & 0.510 & 0.533 & 0.545 & 0.459 & 0.452 \\
                           & AS & 0.591 & 0.311 & 0.549 & 0.609 & 0.581 & 0.516 & 0.551 \\
                           & LUVLi & 0.557 & 0.310 & 0.549 & 0.584 & 0.588 & 0.505 & 0.525 \\
                           & AWing & 0.572 & 0.312 & 0.515 & 0.578 & 0.572 & 0.502 & 0.512 \\
                           & SDFL & 0.576 & 0.315 & 0.550 & 0.585 & 0.583 & 0.504 & 0.515 \\
            ($\uparrow$)   & SDL & 0.589 & 0.315 & 0.566 & 0.595 & 0.604 & 0.524 & 0.533 \\
                           & ADNet & 0.602 & 0.344 & 0.523 & 0.580 & 0.601 & 0.530 & 0.548 \\
                           & SLPT & 0.596 & 0.349 & 0.573 & 0.603 & 0.608 & 0.520 & 0.537 \\
                           & HIH & 0.605 & 0.358 & 0.601 & 0.613 & 0.618 & 0.539 & 0.561 \\
    \cmidrule{2-9}
    & LDEQ & \textbf{0.624} & \textbf{0.373} & \textbf{0.614} & \textbf{0.631} & \textbf{0.631} & \textbf{0.552} & \textbf{0.574}\\
    \bottomrule
  \end{tabular}
  }
  \caption{$\text{AUC}_{10}$ and $\text{FR}_{10}$ on the WFLW test set.}
  \label{tab:WFLW performance AUC FR}
\end{table}

\section{Experiments}

The aim of these experiments is to demonstrate that: 1) LDEQ is a state-of-the-art landmark detection model for high precision settings like faces, and 2) the LDEQ objective can be modified at inference time on videos to include a recurrence loss. This increases temporal coherence without decreasing accuracy, a common pitfall of popular filters. 

\subsection{Landmark accuracy}

\begin{figure*}[t!]
  \centering
   \includegraphics[width=\linewidth]{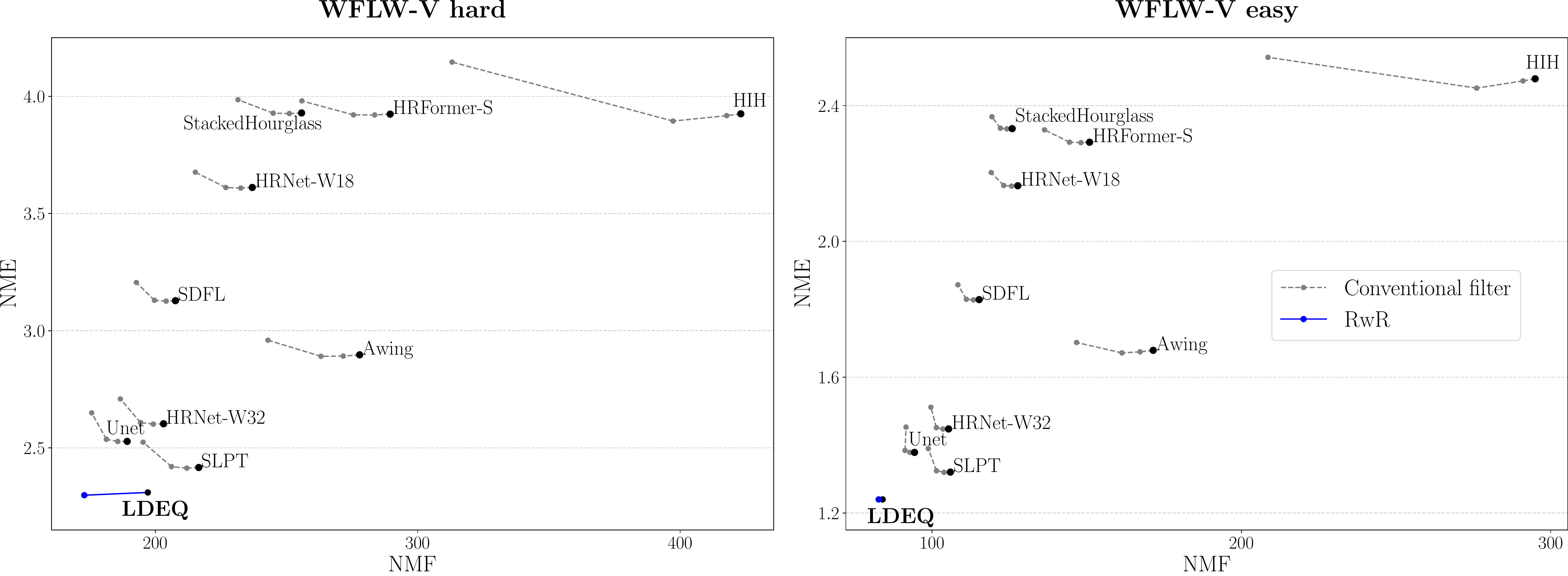}
   \caption{NME (accuracy) vs NMF (temporal coherence) for various models, on the hard (left) and easy (right) WFLW-V subsets. We compare our LDEQ model using our RwR scheme (blue) to several baselines that use an exponential moving average filter with three different window sizes (gray). For hard videos susceptible to flicker, RwR on LDEQ decreases NMF by $12\%$ without increasing NME, contrary to the conventional filter alternative. For easy videos that contain little to no flicker, conventional filters can increase both NME and NMF, while our model converges to the same fixed point with or without RwR. These results are given in tabular form in Appendix C.} 
   \label{fig:NMEvsNMF}
\end{figure*}

We compare the performance of LDEQ to state-of-the-art models on the WFLW dataset \cite{Wayne2018LookAtBoundary}, which is based on the WIDER Face dataset \cite{Yang2016WiderFace} and is made up of $7500$ train images and $2500$ test images. Each image is annotated with a face bounding box and $98$ 2D landmarks. Compared to previous datasets, WFLW uses denser facial landmarks, and introduces much more diversity in poses, expressions, occlusions and image quality (the test set is further divided into subsets reflecting these factors). This makes it uniquely appropriate for training landmark detectors meant to be deployed on real-world data, like the videos of WFLW-V.

The common evaluation metrics for the WFLW test set are the Normalized Mean Error (see \cref{eq: NME}), the Area Under the Curve (AUC), and the Failure Rate (FR). The AUC is computed on the cumulative error distribution curve (CED), which plots the fraction of images with NME less or equal to some cutoff, vs.\ increasing cutoff values. We report $\text{AUC}_{10}$, referring to a maximum NME cutoff of $10$ for the CED curve. Higher AUC is better. The $\text{FR}_X$ metric is equal to the percentage of test images whose NME is larger than $X$. We report $\text{FR}_{10}$; lower is better.

We train our LDEQ for $60$ epochs using the pre-cropped WFLW dataset as per \cite{Lan2021HIH}, and the common data augmentations for face landmarks: rotations, flips, translations, blurs and occlusions. We found the Anderson and fixed-point-iteration solvers to work best over the Broyden solver used in the original DEQ model \cite{Bai2019DEQs, Bai2020MDEQs}. By setting a normalization temperature of $T=2.5$, convergence to fixed points only takes around $5$ solver iterations. The NME, AUC and FR performance of LDEQ can be found in tables \ref{tab:WFLW performance NME} and \ref{tab:WFLW performance AUC FR}. We outperform all existing models on all three evaluation metrics, usually dominating individual subsets as well, which is important when applying LDEQ to video data.

\subsection{Landmark temporal coherence}
\label{subsec: exp-landmark coherence}
We evaluate the performance of LDEQ on the WFLW-V dataset, for both landmark accuracy (NME) and temporal coherence (NMF). We use RwR with early stopping after 2 solver iterations, allowing some adaptive compute (1 solver iteration) in cases where two consecutive frames are almost identical. Our baselines include previous state-of-the-art models that have publicly available weights \cite{Lan2021HIH, Xia2022SLPT, Lin2021StructureCoherent, Wang2019AdaptiveWingLoss}, as well as common architectures of comparable parameter count, which we trained with our own augmentation pipeline. We apply a conventional filtering algorithm, the exponential moving average, to our baselines. This was found to be more robust than more sophisticated filters like Savitzky-Golay filter \cite{SavitzkyGolay1964Filter} and One Euro filter \cite{Casiez2012OneEuroFilter}.

The NME vs.\ NMF results are shown in \cref{fig:NMEvsNMF} for all models, for the hard and easy WFLW-V subsets. For videos that contain little uncertainty in landmarks (WFLW-V easy), there is little flickering and conventional filtering methods can mistakenly smooth out high frequency signal (\eg eyes and mouth moving fast). For videos subject to more flickering (WFLW-hard), these same filtering techniques do indeed improve the NMF metric, but beyond a certain smoothing factor this comes at the cost of increasing the NME. In contrast, LDEQ + RwR correctly smooths out flicker for WFLW-W hard without compromising performance on WFLW-V easy. This improvement comes from the fact that the RwR loss in \cref{eq: relaxed RwR minimization} contains both a smoothing loss plus a log likelihood loss that constrains output landmarks to be plausible solutions, while conventional filters only optimize for the former.

\section{Conclusion}

We adapt Deep Equilibrium Models to landmark detection, and demonstrate that our LDEQ model can reach state-of-the-art accuracy on the challenging WFLW facial dataset. We then bring the attention of the landmark community to a common problem in video applications, whereby landmarks flicker across consecutive frames. We contribute a new dataset and metric to effectively benchmark solutions to that problem. Since DEQs solve for an objective in the forward pass, we propose to change this objective at test time to take into account new information. This new paradigm can be used to tackle the flickering problem, by adding a recurrent loss term at inference that wasn't present at training time (RwR). We show how to solve for this objective cheaply in a way that leads to state-of-the-art video temporal coherence. We hope that our work brings attention to the potential of deep equilibrium models for computer vision applications, and in particular, the ability to add loss terms to the forward pass process, to leverage new information at inference time.

\clearpage
{\small
\bibliographystyle{ieee_fullname}
\bibliography{egbib}
}

\clearpage
\section*{\large{Appendix A: More details on making our WFLW-V dataset}}

In this section we detail the procedure used to collect, label and curate the $1000$ videos that make up the WFLW-V dataset. 

\subsection*{Step 1: Video search}

We start by producing a list of $100$ YouTube search strings, that we think would be correlated with videos conducive to landmark uncertainty. These search strings fall within 7 categories: ``Skin care \& Makeup'' (\eg \textit{how to put lipstick}), ``Hair \& Beard care'' (\eg \textit{how to cut your own hair}), ``Singing \& Podcasts'' (\eg \textit{how to setup your mic}), ``Brass instruments'' (\eg \textit{learn to play the French horn}''), ``Eating'' (\eg \textit{how to eat fast}), ``Smoking'' (\eg \textit{how to smoke the cigar}), and ``Miscellaneous'' (\eg \textit{how to brush your teeth}''). Each English search string is translated to $10$ languages, to produce more diverse videos: French, German, Spanish, Italian, Portuguese, Catalan, Czech, Danish, Estonian, Dutch.

We use YouTube filters to search for videos less than $4$ minutes long, and with a CC BY licence. This licence is the most permissive creator licence. It allows reusers to distribute, remix, adapt the video, and even to use it for commercial use. We only consider videos that have a frame rate between $24$ and $31$ fps inclusive. This is mostly to exclude all videos like kid cartoons that have very low fps. In total, this step produces around $15,000$ videos. 

\subsection*{Step 2: Video cleaning}

Our task is now to find $5$s of contiguous \textit{clean} face for each video. A \textit{clean} face is a real human face, at least $20\%$ visible, from a single person, without video or camera filters (\eg face filters, jump cuts). We also limit the number of videos that come from the same youtuber, so as not to lower diversity. We use the most popular face detector, the Multi-task Cascaded Convolutional Networks (MTCNN) \cite{Zhang2017MTCNN} to help with video cleaning. In total, this leaves around $2,000$ videos. 

\subsection*{Step 3: Video annotation}

We use an oracle made up of $45$ pretrained models, including $15$ large Unets (larger than our LDEQ backbone), $15$ HRNets-W48, and $15$ HRFormer-B. We average the final heatmap of each model to create a mean heatmap, from which we extract our oracle predictions. We found that the bounding box from the MTCNN model are jittery, which in turns facilitates jitter and flicker for landmarks. To fix this we bootstrap our oracle to the bounding box detection. This is done by using the original MTCNN bounding box, finding landmarks, defining a new bounding box based on the smallest/largest landmark coordinates and a scaling margin factor of 1.2, finding landmarks in this new bounding box, and so on. This is repeated for 3 iterations, after which the bounding box values have have converged. We use the landmark predictions on the final bounding box as our oracle predictions. 

As our oracle outputs the mean of an ensemble of $M$ independent models, the error on this mean is given by $\sigma/\sqrt{M}$, where $\sigma$ is the standard deviation of the $M$ predictions. For $M=45$ we measured this error to be $\sim 0.2\%$ of the mean for the hard subset of WFLW-V. 
\begin{figure*}[t!]
  \centering
   \includegraphics[width=0.75\linewidth]{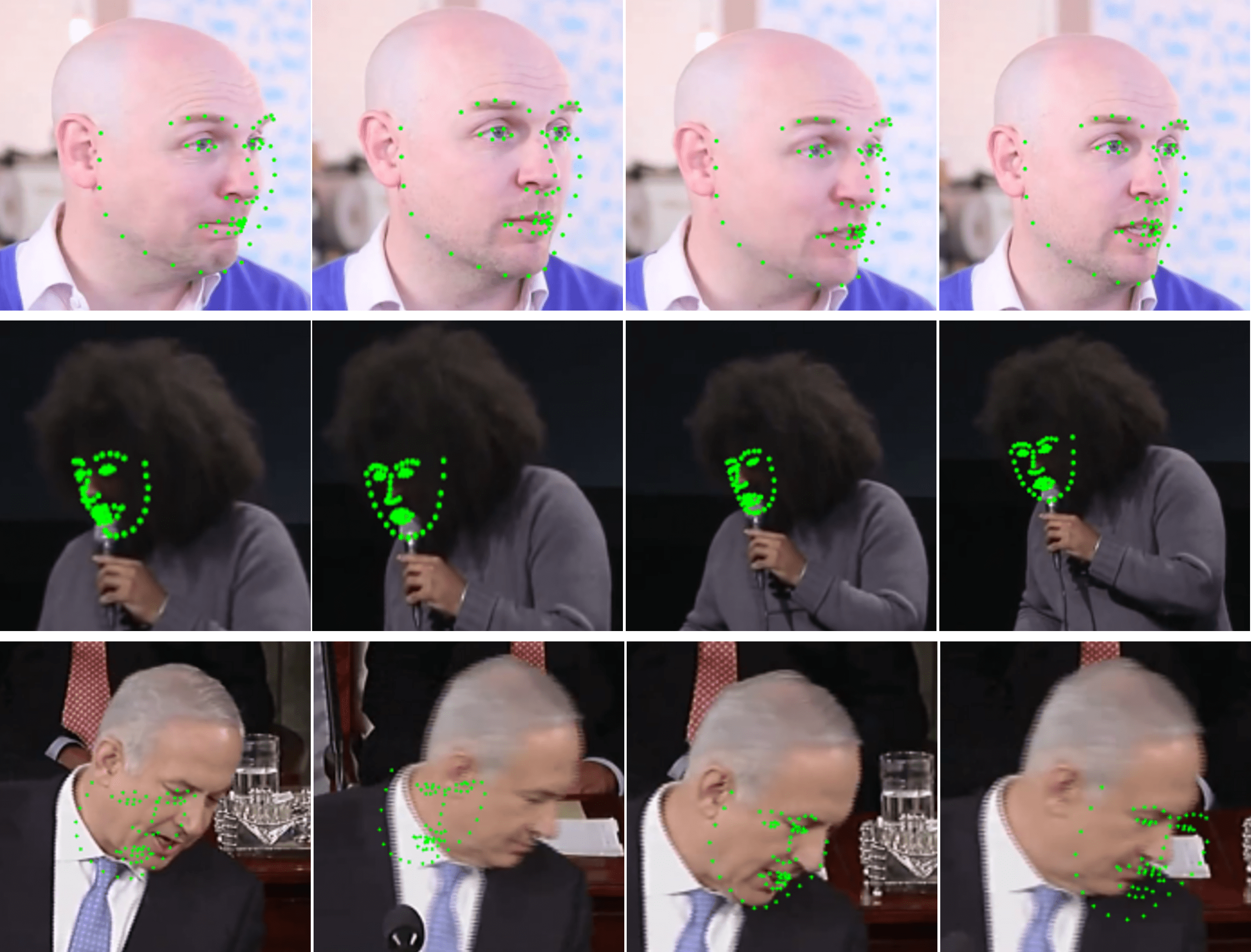}
   \caption{Examples of poorly labelled videos in the 300-VW dataset. We show three levels of labelling errors from top to bottom: medium, bad, very bad. Our new WFLW-V dataset uses much stronger labellers and was checked frame by frame to avoid such errors.}
   \label{fig:examples of 300 VW errors}
\end{figure*}

\subsection*{Step 4: Video verification}

We verify each video frame by frame for issues. In some cases, videos are discarded because the degree of uncertainty or occlusion is too high that even a ``human best guess'' wouldn't be good. This includes videos with very large poses as well. In other cases, particularly for hard videos, some models in the ensemble are visibly mistaken. These models are singled out and removed for problematic frames. These cases are rare ($\sim 25$) but worth correcting so we don't bias the dataset by only including videos where our oracle does best. 

\subsection*{Step 5: Subset creation}

Since we have access to all $45$ model predictions in the ensemble, it is easy to see the average variance of these models for each video. This score correlates well with uncertainty, and we use it to rank all videos from hard to easy. We used the top 500 videos for WFLW-hard, and the bottom 500 for WFLW-easy.

\section*{\large{Appendix B: Errors in 300-VW}}

The 300-VW dataset \cite{Shen2015FirstFacialLandmarkTracking} was labelled using the now obsolete models from \cite{Chrysos2015DeformableFaceTracking} and \cite{Tzimiropoulos2015ProjectOutCascadedRegression}. This results in several labelling errors (\cref{fig:examples of 300 VW errors}) that have gone unnoticed. Errors in the ground truth of datasets lead to misleading insights and models that generalize poorly to real-world settings. 

We also note that many (perhaps all) of the videos in 300-VW do not have a creative commons licence, and so the legality of their use for industrial research labs may be more ambiguous.

\section*{\large{Appendix C: WFLW-V Results}}

We show the results of Figure 5 in tabular form in \cref{tab:WFLW-V Figure 5 in tab form}. We compare our RwR scheme to the exponential moving average (ema), and show that contrary to ema, our method can improve temporal coherence without lowering accuracy. We tried the following ema weights: [0.005, 0.01, 0.02, 0.05, 0.1, 0.15, 0.2, 0.3, 0.4, 0.5, 0.6, 0.7, 0.8, 0.9]. When considering all baselines at once, we found that a weight $0.15$ struck the best balance between lowering NMF without increasing NME too much. This was also better than the grid searched Savitzky-Golay filter \cite{SavitzkyGolay1964Filter} and One Euro filter \cite{Casiez2012OneEuroFilter}. The only exception was the HIH model which is both very jittery and flickery, and for which we used an ema weight of 0.3.

Finally, we found that more augmentations can help performance on WFLW-V while reducing performance on the WFLW test set. This is likely because the WFLW-V dataset is more diverse than the WFLW test set, and unecessary augmentations on WFLW can reduce performance. We therefore retrained LDEQ with more augmentations to get the best performance on WFLW-V.

\begin{table}[h]
  \centering
  \resizebox{\columnwidth}{!}{
  \begin{tabular}{@{}c|cc|cc||cc@{}}
    \toprule
    Method & \multicolumn{2}{c|}{WFLW-V hard} & \multicolumn{2}{c||}{WFLW-V easy} & \multicolumn{2}{c}{WFLW-V FULL} \\
    \cmidrule{2-7}
    ~ & NME & NMF & NME & NMF & NME & NMF \\
    \midrule
HIH & 3.93 & 423.11 & 2.48 & 294.94 & 3.20 & 359.03 \\ 
+ ema & 4.15 & 313.07 & 2.54 & 208.60 & 3.34 & 260.84 \\
StackedHourglass & 3.93 & 255.74 & 2.33 & 125.91 & 3.13 & 190.82 \\ 
+ ema & 3.99 & 231.52 & 2.37 & 119.35 & 3.18 & 175.43 \\
HRFormer-S & 3.92 & 289.50 & 2.29 & 150.91 & 3.11 & 220.21 \\ 
+ ema & 3.98 & 255.85 & 2.33 & 136.37 & 3.15 & 196.11 \\
HRNet-W18 & 3.61 & 236.96 & 2.16 & 127.72 & 2.89 & 182.34 \\ 
+ ema & 3.68 & 215.26 & 2.20 & 119.13 & 2.94 & 167.20 \\
SDFL & 3.13 & 207.68 & 1.83 & 115.22 & 2.48 & 161.45 \\ 
+ ema & 3.21 & 192.77 & 1.87 & 108.35 & 2.54 & 150.56 \\
Awing & 2.90 & 277.86 & 1.68 & 171.48 & 2.29 & 224.67 \\ 
+ ema & 2.96 & 242.95 & 1.70 & 146.71 & 2.33 & 194.83 \\
HRNet-W32 & 2.60 & 203.15 & 1.45 & 105.35 & 2.03 & 154.25 \\ 
+ ema & 2.71 & 186.69 & 1.51 & 99.62 & 2.11 & 143.15 \\
Unet & 2.53 & 189.28 & 1.38 & 94.35 & 1.95 & 141.81 \\ 
+ ema & 2.65 & 175.76 & 1.45 & 91.58 & 2.05 & 133.67 \\
SLPT & 2.42 & 216.56 & 1.32 & 105.93 & 1.87 & 161.25 \\ 
+ ema & 2.52 & 195.35 & 1.39 & 98.79 & 1.96 & 147.07 \\
\bottomrule
LDEQ & 2.31 & 197.16 & 1.24 & 84.03 & 1.77 & 140.59 \\
+ RwR & 2.30 & 172.95 & 1.24 & 82.74 & \textbf{1.77} & \textbf{127.85} \\
    \bottomrule
  \end{tabular}
  }
  \caption{NME and NMF on the WFLW-V dataset, comparing the effect of an exponential moving average smoothing (ema) with our recurrence without recurrence scheme.}
  \label{tab:WFLW-V Figure 5 in tab form}
\end{table}


\end{document}